Predicting the Performance of Minimax and Product
in Game-Tree Searching[1]

Ping-Chung Chi
Dana S. Nau
Computer Science Department
University of Maryland
College Park, Md 20742

ABSTRACT

The discovery that the minimax decision rule performs poorly in some games has sparked interest in possible alternatives to minimax. Until recently, the only games in which minimax was known to perform poorly were games which were mainly of theoretical interest. However, this paper reports results showing poor performance of minimax in a more common game called kalah. For the kalah games tested, a non-minimax decision rule called the product rule performs significantly better than minimax.

This paper also discusses a possible way to predict whether or not minimax will perform well in a game when compared to product. A parameter called the rate of heuristic flaw (rhf) has been found to correlate positively with the performance of product against minimax. Both analytical and experimental results are given that appear to support the predictive power of rhf.

## 1. Introduction

Since the discovery of pathological games [3,4], two questions have attracted a fair amount of research interest. First, is it beneficial to search deeper in various real games? Second, since game tree pathology occurs when using the well known minimax back-up rule in some games, are there alternatives that might do better?

Pearl [7,8] suggested that one should consider the product rule as a way to combine values from an evaluation function. Nau, Purdom, and Tzeng [5] did some experiments and found that in a class of board splitting games, product almost always performed better than minimax and that the product rule avoided pathology. But so far, poor performance of minimax relative to other decision rules has not been observed in games people actually play. By the results of experiments on a more common game called kalah, Slagle and Dixon [10] found that a decision procedure called "M & N" performed significantly better than minimax. However, the M & N rule has a great resemblance to minimax.

Underlying the above questions is a more fundamental issue: Why does minimax perform well in many games, and why does it perform poorly in some others?

Avoidance of sibling dependencies [6] and avoidance of traps [8] have been proposed to be causes of peculiarities such as game tree pathology and the better performance of non-minimax back-up rules in some games. But these two characteristics have more to do with the structure of the game tree itself than they have to do with the heuristics being used.

Abramson's studies [1] on board splitting games showed that one can avoid pathology by improving the heuristics. Apparently this improvement of heuristics can be credited to the existence and detection of "traps."

---

[1] This work was supported in part by an NFS Presidential Young Investigator Award to Dana S. Nau, with matching funds provided by IBM, General Motors, and Martin Marietta; and by NSF grant NSFD CDR-85-00108.

49

This paper explores the issues of trap avoidance and sibling dependencies on the game of kalah and two modifications of kalah. By means of Monte Carlo studies on these games, three interesting results have been observed:

(1) On the average, the product rule performs better than minimax on the kalah games tested.

(2) For the games tested, a parameter called the rate of heuristic flaw appears to be a good predictor of whether minimax will do better than the product rule.

(3) For the games tested, the existence of traps correlates negatively with the performance of minimax playing against product – a result that appears to conflict with what one might predict from Pearl's and Abramson's studies.

## 2. A parameter for game tree measurement

### 2.1. The rate of heuristic flaw

Let G be a zero sum, perfect information game between two players called *max* and *min*, and assume that G has no ties. Let e(.) be a static evaluation function for G. If n is a board position in G, let Win(n) and Loss(n) denote the events that n is a forced win for *max* or forced loss for *max* respectively.

Consider two board positions, m and n. If

$e(m) \geq e(n)$,

then e suggests that m is no worse than n. But an exhaustive search of the game tree may reveal that m is worse than n. More precisely, it may turn out that

Loss(m) and Win(n).

In this case, the evaluation function has failed to give a correct opinion about m and n. An event like this is called a heuristic flaw. The rate of heuristic flaw, denoted by rhf, is defined to be the quantity

$Pr\{Win(n) \mid Loss(m)\ \&\ e(m) \geq e(n)\}$.

In practice, we will usually impose the restriction that m and n are at the same depth of the game tree, or even more strongly, that m and n are siblings. As indicated by the experiments in Section 3, rhf appears to be useful for measuring static evaluators in relation to back-up rules and decision making.

It is apparent that the smaller rhf is, the better e can be expected to be. In the extreme case, if

$Pr\{Win(n) \mid Loss(m)\ \&\ e(m) \geq e(n)\} = 0$,

then e is practically a perfect evaluator, in the sense that a player who uses e will always move to forced win nodes whenever they exist. This is made precise in the following theorem, whose proof appears in [2].

**Theorem 1.** Suppose we have

$Pr\{Win(n) \mid Loss(m)\ \&\ e(m) \geq e(n)\ \&\ Tip(m)\ \&\ Tip(n)\} = 0$,

where Tip(m) denotes the event that m is a tip node of the the search tree for the current board position. Then by moving to the position with the highest minimax value if it is *max*'s move, max will move to a forced win node with probability 1 when it exists. (An analogous statement holds when it is *min*'s move.)

### 2.2. Minimax versus product

Now, let us compare the minimax back-up rule with the product rule. For simplicity, assume that the game tree is binary. Assume further that it is *max*'s move at some node c, and let m and n be the children of c. Then

50

$$\Pr\{\text{Win}(c)\}$$

$$= \Pr\{\text{Win}(m) \text{ or } \text{Win}(n)\}$$

$$= \Pr\{\text{Win}(m)\} + \Pr\{\text{Loss}(m)\} \Pr\{\text{Win}(n) \mid \text{Loss}(m)\}.$$

Thus,

$$\Pr\{\text{Win}(c) \mid e(m) \geq e(n)\}$$

$$= \Pr\{\text{Win}(m) \text{ or } \text{Win}(n) \mid e(m) \geq e(n)\}$$

$$= \Pr\{\text{Win}(m) \mid e(m) \geq e(n)\} + \Pr\{\text{Loss}(m) \mid e(m) \geq e(n)\} \Pr\{\text{Win}(n) \mid \text{Loss}(m) \ \& \ e(m) \geq e(n)\}$$

$$= \Pr\{\text{Win}(m) \mid e(m) \geq e(n)\} + \Pr\{\text{Loss}(m) \mid e(m) \geq e(n)\} \text{ rhf}. \qquad (1)$$

In general, rhf can take on any value between 0 and 1. But when $e(m) \geq e(n)$, Loss(m) presumably makes Win(n) less likely. Thus we can assume

$$0 \leq \Pr\{\text{Win}(n) \mid \text{Loss}(m) \ \& \ e(m) \geq e(n)\} \leq \Pr\{\text{Win}(n) \mid e(m) \geq e(n)\}$$

which is

$$0 \leq \text{rhf} \leq \Pr\{\text{Win}(n) \mid e(m) \geq e(n)\}.$$

Suppose on one extreme that rhf is close to zero. Then from (1) we have

$$\Pr\{\text{Win}(c) \mid e(m) \geq e(n)\} \simeq \Pr\{\text{Win}(m) \mid e(m) \geq e(n)\},$$

which says that the strength of m is a good representation of the the strength of c. This suggests that in this case one would prefer the minimax back-up rule.

The opposite extreme occurs when sibling nodes are independent, i.e., when the event that a node is a win or loss has no effect on whether its siblings are wins or losses. In this case,

$$\text{rhf} = \Pr\{\text{Win}(n) \mid \text{Loss}(m) \ \& \ e(m) \geq e(n)\} = \Pr\{\text{Win}(n) \mid e(m) \geq e(n)\}.$$

Thus, if $e(m)$ and $e(n)$ are good approximations of $\Pr\{\text{Win}(m) \mid e(m) \geq e(n)\}$ and $\Pr\{\text{Win}(n) \mid e(m) \geq e(n)\}$, then from (1),

$$\Pr\{\text{Win}(c) \mid e(m) \geq e(n)\}$$

$$= e(m) + (1 - e(m)) \ e(n)$$

$$= 1 - (1 - e(m)) \ (1 - e(n)),$$

which is precisely the formula for the product rule described in [7,8]. This suggests that in this case the product rule would be preferred.

According to the naive derivations above, we would expect that minimax will perform better compared to product when rhf is small, and that product will perform better when rhf is large. Whether or not this is the case is tested in Section 3 using the game of kalah and two modifications of kalah.

### 3. Simulation results

This section describes the results of Monte Carlo experiments on some classes of kalah games and modified kalah games.

A detailed description of the game of kalah can be found in [9]. For the purpose of understanding the results in this section, the reader should know that kalah is a moderately complex game, perhaps on a par with checkers [9], and that the object of the game is to capture stones from a board. Certain kinds of moves may result in a right to move again for the current player, and other moves may result in a capture of a large number of stones. Also, the game ends prematurely if one player has no move.

By taking away some of these rules one can get several interesting modifications of kalah.

(1) One can allow a game to continue even when one player has no move, by allowing him to skip a move. We call this game "no-premature kalah."

(2) One can disallow go-agains. We call this game "no-go-again kalah."

(3) One could disallow captures. However, we do not consider this game here, because it takes too much computer time to search the entire game tree to get precise measurements of parameters such as the rate of heuristic flaw.

From the analyses in Section 2, we would expect that the larger rhf is, the better product will perform compared with minimax. To test the relevance of this hypothesis, Monte Carlo experiments were carried out for kalah, no-premature kalah, and no-go-again kalah.

Since the winner of a kalah game is the player who gets the most stones, the evaluation function used in these games was the "kalah advantage," i.e., the difference of the numbers of stones acquired by *max* and *min*. This is one of the evaluation functions used by Slagle and Dixon[10].

Using a method for generating random games similar to the method used by Slagle, 1000 game boards were generated. These boards were for kalah games with 3 bottom holes, each hole containing at most 6 stones. They were used as initial boards for all three kinds of kalah games.

In order to estimate rhf in a game, recall that by definition,

$$\text{rhf} = \Pr\{\text{Win}(n) \mid \text{Loss}(m) \ \& \ e(m) \geq e(n)\}$$

$$= \Pr\{\text{Win}(n) \ \& \ \text{Loss}(m) \ \& \ e(m) \geq e(n)\} \ / \ \Pr\{\text{Loss}(m) \ \& \ e(m) \geq e(n)\}.$$

Therefore, rhf can be estimated by estimating $\Pr\{\text{Win}(n) \ \& \ F\}$ and $\Pr\{F\}$ where F is the event "Loss(m) & $e(m) \geq e(n)$."

To estimate $\Pr\{\text{Win}(n) \ \& \ F\}$ and $\Pr\{F\}$, we randomly made 4 moves into the game for each of the 1000 initial boards. Among the children of the boards thus generated, tallies were made for the occurrences of "Win(n) & F" and the occurrences of F. The quotient of these two numbers was used as the estimation of rhf under the restriction that m and n are siblings. The simulation results are given in the following:

| Game Type | Number of Times F Occurs | Number of Times F & Win(n) Occurs | Estimation of rhf |
|---|---|---|---|
| kalah | 905 | 238 | 0.263 |
| no-premature kalah | 947 | 218 | 0.230 |
| no-go-again kalah | 1083 | 320 | 0.295 |

From these data, we are 95% confident (by a one-tailed hypothesis testing) in accepting that rhf is higher in kalah than in no-premature kalah. More careful analyses about the methods used for estimating rhf and interpretations of the results are given in [2].

Of the three types of games, no-go-again kalah has the largest rhf and no-premature kalah has the least rhf. By our hypothesis, we would expect product to perform better on no-go-again kalah and ordinary kalah than on no-premature kalah.

To test whether this is actually the case, the same 1000 boards were used as initial boards for contests between minimax and product at search depths varying from 2 to 7 (for depth 1 both back-up rules are the same). Each initial board was used twice, once with minimax moving first and once with product moving first. From the outcomes of each pair of games thus played, it was determined which player had a better performance. For each pair of games, if one player won both games in the pair, then that player was considered to perform better on that pair. If neither player won both games in the pair, then the pair was not considered.

To save computing time, the computer program did not necessarily examine all 1000 games in every contest. Instead, it stopped whenever at least 100 critical pairs (where the two players performed unevenly) were found and a significant difference in the performance of the two players was reached. The simulation results can be summarized in the following three tables. The significance figures were obtained with one-tailed hypothesis testing, with the null hypothesis that the two players were performing equally well.

Consider the values of the percentage of pairs of games where minimax does better. They should be about 50% if the the null hypothesis is true. Each significance figure in the table gives the probability that the observed deviation from 50% could have arisen by chance if the null hypothesis is true. Thus, the smaller the significance figure is, the less believable is the null hypothesis that minimax and product perform equally well.

Table 1. Results of contests of minimax against product in kalah:

| Search depth | Number of pairs | Number of pairs minimax does better | Percentage of pairs minimax does better | Significance | Conclusion |
|---|---|---|---|---|---|
| 2 | 101 | 27 | 26.7% | <0.1% | product is better |
| 3 | 101 | 66 | 65.3% | 0.1% | minimax is better |
| 4 | 101 | 39 | 38.6% | 1.1% | product is better |
| 5 | 271 | 141 | 52.0% | 25.1% | not significant |
| 6 | 101 | 42 | 41.6% | 4.6% | product is better |
| 7 | 114 | 48 | 42.1% | 4.6% | product is better |

Table 2. Results of contests of minimax against product in no-premature kalah:

| Search depth | Number of pairs | Number of pairs minimax does better | Percentage of pairs minimax does better | Significance | Conclusion |
|---|---|---|---|---|---|
| 2 | 108 | 63 | 58.3% | 4.2% | minimax is better |
| 3 | 101 | 85 | 84.2% | <0.1% | minimax is better |
| 4 | 249 | 136 | 54.6% | 7.2% | not significant |
| 5 | 101 | 80 | 79.2% | <0.1% | minimax is better |
| 6 | 101 | 64 | 63.4% | 1.9% | minimax is better |
| 7 | 101 | 71 | 70.3% | <0.1% | minimax is better |

Table 3. Results of contests of minimax against product in no-go-again kalah:

| Search depth | Number of pairs | Number of pairs minimax does better | Percentage of pairs minimax does better | Significance | Conclusion |
|---|---|---|---|---|---|
| 2 | 101 | 32 | 31.7% | <0.1% | product is better |
| 3 | 103 | 43 | 41.7% | 4.7% | product is better |
| 4 | 105 | 44 | 41.9% | 4.9% | product is better |
| 5 | 114 | 48 | 42.1% | 4.6% | product is better |
| 6 | 101 | 34 | 33.7% | <0.1% | product is better |
| 7 | 101 | 40 | 39.6% | 1.8% | product is better |

Some interesting observations can be made on the data shown above:

(1) At a confidence level of 95%, we accept that product performs better than minimax at search depths 2, 4, 6, and 7 in the game of kalah. Only at depth 3 is product significantly worse than minimax. So, product appears to be superior to minimax at most search depths

53

in this class of kalah games.

(2) At a confidence level of 95%, we accept that minimax performs better at every search depth except 4 in no-premature kalah.

(3) At the same confidence level, we accept that product performs better at every search depth in no-go-again kalah.

(4) The performance of product against minimax correlates positively with the rhf values for the three types of kalah games – a result that matches our prediction about rhf values.

(5) Another interesting observation is that the performance of product relative to minimax changes dramatically from ordinary kalah to no-premature kalah. The only causative factor that we can single out at this point is the existence of premature endings of ordinary kalah as compared to no-premature kalah. It appears that the existence of premature endings in this case contributed to a higher rhf value and also made product to perform better relative to minimax.

## 4. Conclusions and further speculations

The results of our study are summarized below:

(1) Product plays better than minimax at most search depths in the game of kalah with three bottom holes. And it appears that product performs even better in no-go-again kalah. This is a remarkable result, since it is the first time product has been found to be better than minimax in a game people play.

(2) In our experiments, the rate of heuristic flaw appears to be a good predictor of how well minimax will perform in comparison to product. One could argue that for most real games it may be computationally intractable to measure this parameter, since one would have to search the entire game tree. But one can generally make intuitive estimates of rhf without searching the entire game tree.

For example, consider the game of chess. Imagine two game boards, A and B, which are identical except that in board A, White has one more knight than in B. One would usually be confident in saying that if A is a forced loss for White, B will also be a forced loss for White. So, no matter what value is given to the knight by the evaluation function, the ability to tell that A is better than B in this case certainly would make rhf smaller.

(3) From Pearl's and Abramson's studies, one would expect that by introducing traps into a game, one may improve the performance of minimax compared to product. But our results in kalah and modifications of kalah seemed to be the other way around. In our results, the better performance of product was correlated with the higher value of rhf, which might be credited to the existence of premature endings in ordinary kalah as compared to no-premature kalah.

Our results match the naive arguments of Section 2.2 surprisingly well. This may happen partly because the branching factors of the game trees simulated are usually 2 or 3, and Section 2.2 assumed a branching factor of 2. More studies need to be done for higher branching factors. The following questions still are open: Are there other accessible and relevant parameters that could be used to predict the performance of minimax and other decision rules? How might these predictors be related to game tree pathology? What will happen if one uses decision rules other than minimax or product?

The studies of this paper suggest the possibility of answering these and other questions by a combination of intuitive arguments, Monte Carlo experiments and variations of game playing rules.

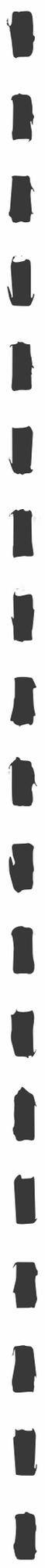